\documentclass[10pt,twocolumn,letterpaper]{article}

\usepackage{cvpr}              % To produce the CAMERA-READY version
\usepackage{graphicx}
\usepackage{amsmath}
\usepackage{amssymb}
\usepackage{booktabs}
\usepackage{multirow}
\usepackage{graphicx} 
\usepackage[labelfont=bf]{caption}

\newcommand\blfootnote[1]{%
  \begingroup
  \renewcommand\thefootnote{}\footnote{#1}%
  \addtocounter{footnote}{-1}%
  \endgroup
}

\usepackage[pagebackref,breaklinks,colorlinks]{hyperref}

\usepackage[capitalize]{cleveref}
\crefname{section}{Sec.}{Secs.}
\Crefname{section}{Section}{Sections}
\Crefname{table}{Table}{Tables}
\crefname{table}{Tab.}{Tabs.}

\def\cvprPaperID{*****} % *** Enter the CVPR Paper ID here
\def\confName{CVPR}
\def\confYear{2022}

\begin{document}

% \title{\LaTeX\ Author Guidelines for \confName~Proceedings}
% \title{Comparison of Humans and Machines on Extreme Image Transforms}
\title{Robustness of Humans and Machines on Object Recognition \\ with Extreme Image Transformations}

\author{Dakarai Crowder $^*$\\
Northeastern University\\
{\tt\small crowder.d@northeastern.edu}
\and
Girik Malik $^{* \dagger}$\\
Northeastern University\\
{\tt\small malik.gi@northeastern.edu}
}
\maketitle
\def\thefootnote{$*$}\footnotetext{These authors contributed equally to this work} \def\thefootnote{\arabic{footnote}}
\blfootnote{$\dagger$ Corresponding author \\ GM is also affiliated with Labrynthe Pvt. Ltd., New Delhi, India}

%%%%%%%%% ABSTRACT
\begin{abstract}
   Recent neural network architectures have claimed to explain data from the human visual cortex. Their demonstrated performance is however still limited by the dependence on exploiting low-level features for solving visual tasks. This strategy limits their performance in case of out-of-distribution/adversarial data. Humans, meanwhile learn abstract concepts and are mostly unaffected by even extreme image distortions. Humans and networks employ strikingly different strategies to solve visual tasks. To probe this, we introduce a novel set of image transforms and evaluate humans and networks on an object recognition task. We found performance for a few common networks quickly decreases while humans are able to recognize objects with a high accuracy.
\end{abstract}

%%%%%%%%% BODY TEXT
\section{Introduction}
\label{sec:intro}
% basic motivation
% talk about the meaning of the object for humans and machines
% link with Ullman's PNAS paper, learning with noise and the use of additional blocks to simulate V1

Driving in heavy rain, snow, or other adversarial conditions, humans rely on their ability to quickly recognize the vehicle or humans in front based on very few high-level cues. These cues indicate the presence of an ``object" instead of trying to accurately predict the object class based on low-level features (vehicle make or color of human's clothes). They use previously learned abstract knowledge of the object and make decisions based on their understanding of how objects interact with the environment. Object recognition is one of the most fundamental problems solved by primates for their everyday functioning and survival. The primate visual system is known to be robust to small perturbations in the scene~\cite{zhou2019humans, eidolons}, and uses fairly sophisticated strategies to recognize objects in adversarial conditions with a very high accuracy. 

Neural networks largely disregard entity-level recognition and ``learn" to recognize objects based on bottom-up cues, like contours, color, textures, etc., that allow them to easily exploit ``shortcuts" in input distribution~\cite{geirhos2018generalisation}. This in-turn affects the performance of networks when attacked with adversarial input. To probe the limits of this gap in human and network performance, we introduce novel image transformations that go beyond the currently employed techniques of adversarial attacks. We test the limits to which networks can perform an object recognition task with our image transforms on CIFAR100~\cite{cifar} dataset. We further test human subjects and networks on an object recognition task with the same image and transform combination, and show humans are more robust in solving the task with much greater accuracy.

\noindent\textbf{Related work } Ullman et al.~\cite{ullman2016atoms} use ``minimal recognizable images" to test the limits of network performance on object recognition, and show that networks are susceptible to even minute perturbations at that level. Rusak et al.~\cite{rusak2020simple} show adversarially trained object recognition model against locally correlated noise improves performance.  By removing texture information and altering silhouette contours, Baker et al.~\cite{baker2018deep} show that networks focus on local shape features.  
% Baradad et al.~\cite{learningtoseebylookingatnoise} try to learn robust visual representations by generating models of noise closer to the distribution of real images.

\noindent\textbf{Contributions:} Humans show robustness to extreme image transforms on object recognition tasks.
\begin{itemize}
    \vspace{-2mm}
    \item We introduce novel image transforms to simulate extreme adversarial attacks  
    \vspace{-2mm}
    \item We present an extensive study of how network performance differs with changes in our transforms
    \vspace{-2mm}
    \item We probe the limit of robustness between humans and networks on object recognition tasks through our transforms
\end{itemize}

% Currently  many neural networks are being built in a way that tries to mirror the function of the primate visual cortex, in order to get a better understanding of the functionalities. So we wanted to see if there were correlations between how a model identifies and object and how a human does. Ullman found that when minute changes are made to an image the ability for models to identify what the object is drops significantly, while humans are still able to identify what that object is. Thus, coming to the conclusion that the current models are not able to recognize images at the scale of a human due to models losing the identifying features they use to differentiate between objects when minimal images are used{*** cite atoms of rec ***}. Bowers et al. also explains that although DNNs are most accurate in classifying images and predicting human error, having high performance on brain and behavioral benchmarks may share on little intersection with the function of biological vision{*** cite deep problems with nn models of human vision ***}. However, VOneNet was designed based off of area v1 in the brain{}. This model has been claimed to have similar functions as area V1 in the brain. 

%------------------------------------------------------------------------
\section{Image Transformations}
\label{sec:transformations}

\begin{figure}
  \centering \includegraphics[width=.478\textwidth]{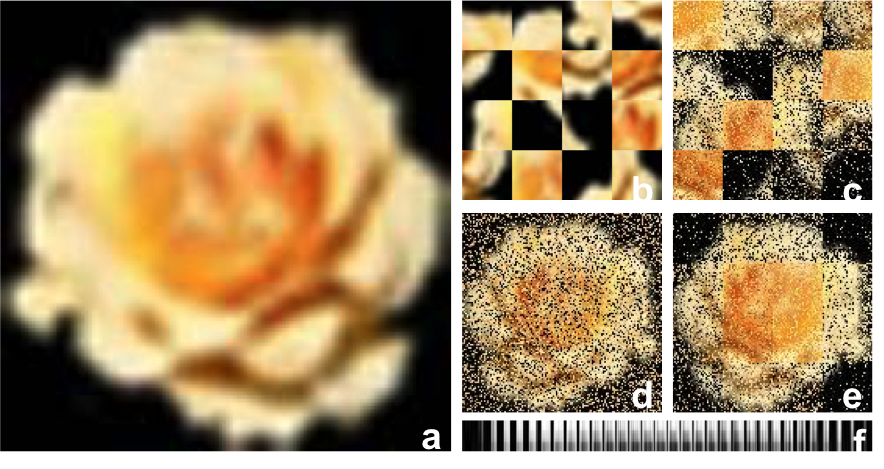}
  \caption{Transforms applied to a CIFAR100 image. (a) non-transformed image, (b) Grid Shuffle with block size 8x8, (c) Local Grid Shuffle with block size 8x8 and probability 0.5, (d) Randomized Image Shuffle with probability 0.5, (e) Within Grid Shuffle with block size 8x8 and probability 0.5, and (f) Color Flatten.}
  \label{fig:fig1}
\end{figure}

To test the limits of human and machine vision on object recognition task with distorted image structures, we introduce five novel image transformations (Figure \ref{fig:fig1}).

\noindent\textbf{Grid Shuffle} divides the image into blocks of equally sized squares, or rectangles. The divided units are individually shuffled with a specified probability (range: [0.0-1.0]).

\noindent\textbf{Randomized Image shuffle} shuffles the pixels within the image based on a specified probability (range: [0.0-1.0]), disregarding any underlying structural properties of the image. For a shuffle probability of 0.5, each pixel’s location has a 50\% chance of being shuffled, while a shuffle probability of 1.0 moves every pixel around, with the image looking like random noise. 

\noindent\textbf{Within Grid Shuffle} divides the image into blocks (similar to Grid Shuffle), but does not shuffle the blocks. Instead, it shuffles the pixels within the blocks with a specified probability. Pixel shuffling within the block is identical to Randomized Image Shuffle, considering each unit in the block to be an individual image.

\noindent\textbf{Local Grid Shuffle} is a combination of Within Shuffle and Grid Shuffle. It divides the image into blocks (like Grid Shuffle), shuffles the pixels within the blocks (like Randomized Image Shuffle), and further shuffles the positions of the blocks. It alters both global and local structure of the image. 

\noindent\textbf{Color Flatten} separates the three RGB channels of the image and flattens the image pixels from 2-dimensional \text{$N \times N$} to three channel separated 1-dimensional vectors of length \text{$N*N$} in row-major order.

\section{Model Selection}
\label{sec:models}
We selected ResNet50, ResNet101~\cite{resnet}, and VOneResNet50~\cite{vonenet} for our experiments with baseline (no shuffle) and image transformations.
VOneResNet50 was selected for its claims of increasing robustness of CNN backbones to adversarial attacks by preprocessing the inputs with a VOneBlock -- mathematically parameterized gabor filter bank -- inspired by the Linear-Nonlinear-Poisson model of V1. It also had a better V1 explained variance on brain-score~\cite{brainscore} benchmark at the time of our experiments. 
We chose ResNet50 since it formed the CNN backbone of VOneResNet50 and we wanted to test the contribution of non-V1-optimized part of VOneResNet50.
We chose ResNet101 for its high average score on brain-score in terms of popular off-the-shelf models that are widely in use.

%------------------------------------------------------------------------

\section{Experiments}
\label{sec:experiments}
%setup/dataset
% why cifar, what experiments

\noindent\textbf{Setup } We evaluated ResNet50, ResNet101, and VOneResNet50 on CIFAR100 against baseline images (without transforms), four shuffle transforms, and Color Flatten transform described in \ref{sec:transformations}. We used the default train-test split of 50,000 training images and 10,000 test images from the dataset, distributed over 100 classes. Each image has 3 channels and $32 \times 32$ pixels. We did not separate a validation set for network fine-tuning to mimic how humans only see a small subset of objects and then recognize them in the wild, without fine-tuning their internal representations.

% justify the use of CIFAR100 as dataset

%training
\noindent\textbf{Training } We trained each model on the baseline and transformed images using the default hyperparameters listed in the repositories of the respective models. For the Grid Shuffle transformation, we used three grid sizes -- $4 \times 4$, $8 \times 8$, and $16 \times 16$ (dividing image into 4 blocks). For Within Grid Shuffle and Local Grid Shuffle we used a combination of three block sizes ($4 \times 4$, $8 \times 8$, and $16 \times 16$) with a shuffle probability of 0.5 and 1.0 for each block. For Randomized Image Shuffle we used shuffle probabilities of 0.5 and 1.0. We used a total of 18 different transformations. 

%For all transforms involving a grid, we limited ourselves to perfect squares given the square shape of the image and ease of handling with humans and networks. *** limitations

%histogram
In the Color Flatten transform, we separated the image channels and flattened the 2D array to 1D in row-major order. An additional Conv1D layer was added to the networks to process the 1D data, feeding it to the individual networks. All networks were trained end-to-end using only the respective transform (and its hyperparameters), without sharing any hyperparameters across same or different types of transforms.

%human trial -- to be done, describe the process of trials -- warmup, test; how the system was setup, etc. Put extra details to supplementary
\noindent\textbf{Human experiments } The authors (D.C. and G.M.) also tested themselves on images randomly sampled from the test set of CIFAR100. Images shown to human participants were scaled to $128 \times 128$ pixel resolution. The trial screen showed an image and a set of 5 possible object classes that could be present in the image, and asked humans to select the object category closest to the one they see in the image. Each participant was trained on 17 images before the start of the experiment. The training trials consisted of images randomly sampled from CIFAR100 dataset, and then a shuffle transform applied with a unique hyperparameter of grid size and shuffle probability. The hyperparameter combination was not repeated during training (see \S\ref{sec:human_setup} for details). 

We evaluated humans and networks on 93 trials during testing. Similar to training phase, we showed a single image during a trial, with or without the transforms. The transforms had same hyperparameters as used with networks. Each transform had about 5 different classes of images sampled from CIFAR100, with every transform applied on every image, showing the same transform multiple times, but not with the same image. We additionally asked humans to indicate their level of confidence for their response. % with their choice of object category in the image.

% move more details to supplementary
% For every transformation not including the color histogram but including a baseline, five images were selected from the validation data set. These images were then uploaded to MTurk. In addition, for each transformation an image was selected for a practice round for the human trials. Each person was asked to select one option out of five classes displayed. They were then asked to give a confidence level allowing the knowledge of a complete guess. The models were also tested against the same images as the human trials for analysis.
% explain 5 different images for each of the 4 different transformations

%limitations -- to be moved to the end of the paper alongside discussions
% There are a few limitations for the human trials. There were only two subjects that these images were tested on. The two subject also know about each transformation. Since the same image was used but had differing probability and box sizes, the subject could start to remember by the colors what an object was.

%\textbf{ResNet50} & \textbf{ResNet101} & \textbf{VOne}                \\ cline(4-6)
%------------------------------------------------------------------------
\section{Results}

\begin{table}[ht]
\caption{Test accuracy for models trained on CIFAR100 dataset}
\label{table:table1}
\begin{tabular}{|@{}p{1.4cm}|p{.3cm}|p{.8cm}|c|c|c|}
\hline

\multirow{2}{*}{\textbf{Transform}}  & \multirow{2}{*}{\textbf{P}} & \multirow{2}{2em}{\textbf{Grid Size}} & \multicolumn{3}{|c|}{\textbf{Accuracy (in \%)}}\\ \cline{4-6}
& & &\textbf{ResNet50} & \textbf{ResNet101} & \textbf{VOne}  \\ \hline

                                                %   p                      gridsize             50        101      vone
\textbf{Baseline}                                   &                      &                    & 47.80    & 72.10    & 99.66  \\ \hline

\multirow{2}{5em}{\textbf{Random Shuffle}}             & 0.5                  &\multirow{2}{*}{}   & 36.56  & 37.53 & 35.08  \\ \cline{2-2} \cline{4-6} 
                                                    & 1                    &                    & 18.80   & 18.79 & 9.81   \\ \hline
\multirow{3}{5em}{\textbf{Grid Shuffle}}            &\multirow{3}{*}{}     & 4x4                & 29.35  & 29.86 & 16.75  \\ \cline{3-6} 
                                                    &                      & 8x8                & 35.79  & 36.69 & 26.05  \\ \cline{3-6} 
                                                    &                      & 16x16              & 44.88  & 46.40  & 36.29  \\ \hline
\multirow{6}{5em}{\textbf{Within Grid Shuffle}}     & \multirow{3}{*}{0.5} & 4x4                & 44.89  & 44.86 & 39.24  \\ \cline{3-6} 
                                                    &                      & 8x8                & 40.14  & 40.11 & 37.47  \\ \cline{3-6} 
                                                    &                      & 16x16              & 37.91  & 38.48 & 36.97  \\ \cline{2-6} 
                                                    & \multirow{3}{*}{1}   & 4x4                & 42.68  & 42.98 & 37.70   \\ \cline{3-6} 
                                                    &                      & 8x8                & 34.73  & 35.29 & 31.59  \\ \cline{3-6} 
                                                    &                      & 16x16              & 25.51  & 25.89 & 17.70   \\ \hline
\multirow{6}{6em}{\textbf{Local Grid Shuffle}} & \multirow{3}{*}{0.5} & 4x4                & 23.94  & 24.40  & 18.66  \\ \cline{3-6} 
                                                    &                      & 8x8                & 26.99  & 26.80  & 21.87  \\ \cline{3-6} 
                                                    &                      & 16x16              & 32.30   & 32.50  & 28.02  \\ \cline{2-6} 
                                                    & \multirow{3}{*}{1}   & 4x4                & 23.03  & 23.42 & 16.64  \\ \cline{3-6} 
                                                    &                      & 8x8                & 24.29  & 24.49 & 19.17  \\ \cline{3-6} 
                                                    &                      & 16x16              & 22.99  & 25.50  & 16.16  \\ \hline
\textbf{Color Flatten}                              &                      &                    & 46.72  & 47.48 & 44.22  \\ \hline
\end{tabular}
\end{table}

VOneResNet50 performs the best on baseline (without transforms) CIFAR100 test images, with 99.6\% accuracy, followed by ResNet101 and ResNet50 (Table \ref{table:table1}). Transforming images significantly reduces performance, with none of the networks being able to reach 50\% accuracy. However, ResNet50 and ResNet101 consistently perform better than VOneResNet50 by a large margin. ResNet101 performs better than ResNet50, but the difference is not significant. In case of Randomized Image Shuffle, the accuracy expectedly decreases with an increase in shuffle probability. For Grid Shuffle, the accuracy monotonically increases with an increase in grid size.

Within Grid Shuffle preserves the local structure of the image, adversarially affecting the performance with an increase in grid size and shuffle probability. Smallest grid size is the least affected by increase in shuffle probability. All networks struggle the most on Local Grid Shuffle. Large grid size with low shuffle probability gives a higher accuracy than other variants of grid and shuffle probability, but is overall lower than the other transforms. Flattening the images across color channels gives a higher accuracy across all transforms. 

\noindent\textbf{Comparison with Human responses } Human subjects performed better than all networks on all but Color Flatten transform (Table \ref{table:table2}). Humans had a perfect accuracy on baseline, Randomized Image Shuffle (0.5 shuffle probability), and $16 \times 16$ Grid Shuffle. 
We ran Ordinary Least Squares regression (OLS) to check if the data from human responses fits the responses from the three networks on the same images shown to the human observers (see Table \ref{tab:summary_table} for comparison of human and network responses on different transforms). Comparing human and network responses for all transforms and their respective hyperparameters, we can reject our null hypothesis that human and machine responses are correlated (Table \ref{tab:ols_alldata}).

We also used OLS on the human and machine responses for individual transforms (Tables \ref{tab:ols_baseline}, \ref{tab:ols_randomshuffle}, \ref{tab:ols_gridshuffle}, \ref{tab:ols_withingridshuffle}, \ref{tab:ols_localshuffle}, \ref{tab:ols_colordist}). For  Grid Shuffle, Local Grid Shuffle, Randomized Image Shuffle, and Within Grid Shuffle, we can safely reject the null hypothesis with 99\% confidence interval. For baseline and Randomized Image Shuffle, human responses might be correlated with all networks. For Color Flatten, we could only reject the null hypothesis with 95\% CI. Few transforms with $16\times 16$ grid might make it easy to understand the basic structure of objects from shape and contours. To test if it aids networks as much as humans, we compared human and network responses only for images that were transformed with a $16\times 16$ grid -- namely Grid Shuffle, Within Grid Shuffle and Local Grid Shuffle -- with both shuffle probabilities (Table \ref{tab:ols_16x16}). We found that we can reject the null hypothesis, concluding that humans employ different techniques than machines to solve seemingly easy tasks.
% should we discuss any ways that we think the model may be identifying objects?  -- i already wrote that in discussions.

\section{Discussion and Conclusion}
% summarize the main results and contributions
% how does it link with human vision
% what might be the causes of discrepancies in the results and what can be concluded from them.
% talk about the claims of VOneNet simulating V1 in their block. Refute gracefully.
% link with the histogram image if possible to talk about how the color intensity distribution might affect performance (weak argument)?
VOneResNet50, ResNet101 and ResNet50 struggle to solve object recognition task on CIFAR100 when trained and tested with our novel transforms. Humans, however, can recognise objects on extreme transforms, implying that humans and networks use very different strategies for object recognition with perturbed images. We show that these differences in strategies are statistically significant. 

Unlike initial layers of CNNs that learn edges, contours, and textures, humans rely on an abstract concept of ``object" representation and further add individual features linking it with object's category. The abstract concept of object helps humans to learn about the characteristics of the object and link it with the accompanying information about the environment, dimensionality, etc.. This happens at various levels of the human visual system~\cite{carandini2005we}. During object's interaction with the environment, humans treat the object (independent of the class) as a whole individual entity, as opposed to parts of it interacting separately. (The representation of the object being referred to here is atomic in nature. For entities with moving parts, individual parts must be treated as individual objects.)  

While CNNs learn representations in a feature specific manner, largely discounting the characteristic properties of the underlying object, humans try to learn the knowledge of features, building on top of objects~\cite{peronamalikscale, koenderink1984structure, koenderink2021structure}. We speculate this might be reflected in our results as well. Networks struggle to link features to an object class when the features are shuffled, but humans' internal representations are robust to the effects of such distortions, as we show in Table \ref{table:table2}. Networks are unable to perform on our easiest transforms with a $16 \times 16$ grid, specifically the $16\times 16$ Grid Shuffle -- which divides the image into 4 blocks and shuffles them, preserving features closest to the baseline images -- demonstrating heavy reliance on features like object shape.

Recent work has shown the ability of CNNs to explain variance in V1, and has gone a step further to create network layers in-line with the variance~\cite{vonenet}. Other works have demonstrated learning with noise~\cite{learningtoseebylookingatnoise}. These networks perform well on adversarial tasks with controlled attacks. We show how they behave when the control is taken away. We think our work can pave the way for adversarial testing with extreme attacks. We are unsure if CNNs are the right contenders for dealing with such attacks, given the primate visual system has dependency on feedback. Recurrent networks have previously demonstrated good performance~\cite{NEURIPS2021linsley} on tasks where CNNs struggle~\cite{pathtracker}. 

We demonstrate that human visual system employs a more robust strategy to solve the object recognition task, and remains largely unaffected by the seemingly extreme changes in images. Our novel transforms highlight a blind spot for the controlled training of adversarial networks. We hope these transforms can help with development/training of robust architectures simulating tolerance of primate visual system to deal with extreme changes in visual scenes often found in everyday settings.

% can also talk about how networks have taken away appearance from tracking and we envision similar robustness for this field.

\section{Limitation and Future Work}

% Limitations 
We used CIFAR100 with 60,000 images distributed over 100 classes. Using a larger dataset could have yielded more stable results. Due to resource limitations, we could collect data from only authors as human subjects. While the network experiments and human data collection were weeks apart, it could still have introduced some bias. We used only 5 unique images across transforms (with its hyperparameters). The tradeoff we faced was between reducing the time spent on human trials to not loose attention v/s using a larger set of images that might have stabilized the results even more. While the classes were unique and the order of presentation was randomly shuffled for each subject, we cannot rule out the possibility of memorization across different hyperparamaters of the transformations.

% future work 
%using coarse dataset like what was done in  human uncertainty paper, 
% Being able to see if the model could become more robust if using the coarse labels that CIFAR100 has, or using soft labels as done in Human uncertainty makes classification more robust, is something that could be explored deeper. This could give some more insight on the affect of soft labels verses fine labels on the models robustness. Transformations made on a annotated image level, could possible give some more inset on what short cuts neural networks may be taking in order to identify an object.
Using coarse and fine labels of CIFAR100, we aim to train these networks on a mix of labels to see if the network learns the representation of an abstract entity like 4-legged animal and could then fine-tune for cat, dog, etc. While some previous work \cite{human_uncertainity} tries to adopt similar approach with human labels, we think the community could benefit with work using extreme transforms.

\section*{Acknowledgements}
\noindent We would like to thank Prof. Ennio Mingolla for inputs at various phases of this work. GM would also like to thank Sobia Shadbar for discussions and inputs on statistical analysis.

%%%%%%%%% REFERENCES
{\small
\bibliographystyle{ieee_fullname}
\bibliography{egbib}
}

%%%%%%%%%% Merge with supplementary materials %%%%%%%%%%
% \widetext
\clearpage
\setcounter{figure}{0}
\setcounter{table}{0}
\setcounter{section}{0}

\makeatletter 
\renewcommand{\thesection}{S\@arabic\c@section}
\renewcommand{\thefigure}{S\@arabic\c@figure}
\renewcommand{\thetable}{S\@arabic\c@table}
\makeatother

% \title{Supplementary Materials}
% \noindent{\Large \textbf{The Challenge of Appearance-Free Object Tracking with Feedforward Neural Networks \\ \\ -- Supplementary Information --}}
\twocolumn[  
    \begin{@twocolumnfalse}
        \begin{center}

             \Large \textbf{Robustness of Humans and Machines on Object Recognition \\ with Extreme Image Transformations \\-- Supplementary Information --}

         \end{center}
     \end{@twocolumnfalse}
]

\section{Human Experiments Setup}
\label{sec:human_setup}
The authors, D.C and G.M. were the two participants in the study for collecting human responses for the different transformations. The experiment was not time bound and could be completed at participants own pace. The experiment was designed to take an average of 20-25 minutes.  We recorded the reaction time for all trials. After every trial, participants were redirected to a screen confirming their submission. They could continue by clicking the ``Continue" button or pressing the spacebar. They were automatically redirected from the confirmation screen to the next screen in 3000ms. We also showed a ``rest screen" after completion of 10 trials, with a progress bar. The rest screen was shown only during main trials and not during practice trials. The time on rest screen was not recorded. 

\paragraph{Experiment design} At the beginning of the experiment, the participants were shown an information screen guiding them about the significance of the experiment and what needs to be done. They could then click ``Continue", which showed an instruction modal pop-up with instructions about what to do. They could view this instruction modal anytime during the experiment by clicking the button ``Instructions" in the top right corner of their screens. 

Participants were shown an image (baseline or transformed) along with five object classes from the CIFAR100 labels. They were asked to identify the object in the image and select the option closest to what they thought the object in the image was. They were also asked to rate their level of confidence on a scale of 1 through 5, ranging from least to most confident. They were given a feedback on their response in the form of correct or incorrect during practice trials but not during the main test trials. Each trial screen also had a short excerpt of instructions. Participants were shown a total of 17 practice trials and 88 test trials.

\paragraph{Software setup} The experiment used Python Flask for backend scripts and logic, and HTML, Bootstrap CSS framework and JavaScript for frontend. The form submission through keys and automatic redirections were done using jQuery on the user side. The server was run on HP Z200 workstation using 1 Intel(R) Xeon(R) CPU and 16 GB RAM.

CIFAR100 has 3-channel images of $32\times 32$ pixel resolution. For showing to human participants, the images were sampled from test set at the original resolution and then scaled to $128\times 128$ pixel resolution using linear interpolation, before applying the required transform on them.

\section{Statistical Analysis of Human and Network Data}
To test the significance of human responses to the data returned from the network, we used Ordinary Least Squares regression (OLS) to fit human and machine data. We used \texttt{statsmodels} library in python for our analysis.

\begin{table*}
\caption{Summary table for OLS Linear Probit Models across different transforms}
\label{tab:summary_table}
\begin{center}
% \begin{tabular}{lcccccccc}
\begin{tabular}{p{0.12\linewidth} p{0.095\linewidth} p{0.095\linewidth} p{0.095\linewidth} p{0.095\linewidth} p{0.095\linewidth} p{0.095\linewidth} p{0.095\linewidth} p{0.095\linewidth}}
\hline
      &    All data     &    Baseline   &   Randomized Image Shuffle   &   Grid Shuffle   &  Within Grid Shuffle  &  Local Grid Shuffle  & Color Flatten & All $16\times 16$ transforms  \\
\midrule
Humans (con.) & 0.7966***  & 1.0000*** & 0.7500*** & 1.0000***  & 0.8667***  & 0.7333***  & 0.0000    & 0.8000***   \\
      & (0.0422)   & (0.2000)  & (0.1830)  & (0.1024)   & (0.0745)   & (0.0889)   & (0.2000)  & (0.0856)    \\
VOneResNet50    & -0.4492*** & -0.4000   & -0.2500   & -0.7333*** & -0.5667*** & -0.3333*** & 0.6000**  & -0.5600***  \\
      & (0.0597)   & (0.2828)  & (0.2588)  & (0.1447)   & (0.1053)   & (0.1257)   & (0.2828)  & (0.1211)    \\
ResNet101    & -0.4576*** & -0.4000   & -0.2500   & -0.7333*** & -0.7000*** & -0.2667**  & 0.6000**  & -0.5200***  \\
      & (0.0597)   & (0.2828)  & (0.2588)  & (0.1447)   & (0.1053)   & (0.1257)   & (0.2828)  & (0.1211)    \\
ResNet50    & -0.4661*** & -0.2000   & -0.2500   & -0.7333*** & -0.6333*** & -0.3667*** & 0.8000**  & -0.6000***  \\
      & (0.0597)   & (0.2828)  & (0.2588)  & (0.1447)   & (0.1053)   & (0.1257)   & (0.2828)  & (0.1211)    \\
\hline
\end{tabular}
\end{center}
Standard errors in parentheses. \\
* p$<$.1, ** p$<$.05, ***p$<$.01
\end{table*}

% \textbf{Model and human accuracy}
\begin{center}
\begin{table*}
\captionsetup{singlelinecheck = false, format= hang, justification=raggedright, font=footnotesize, labelsep=space}
\caption{Model and human accuracy on the same CIFAR100 images}
\label{table:table2}
\begin{tabular}{|l|c|l|r|r|r|r|}
\hline
\multirow{2}{*}{\textbf{Transform}}  & \multirow{2}{*}{\textbf{P}} & \multirow{2}{*}{\textbf{Grid size}} & \multicolumn{4}{|c|}{\textbf{Accuracy (in \%)}}\\ \cline{4-7}
& & &\textbf{ResNet50} & \textbf{ResNet101} & \textbf{VOne}  & \textbf{Human}\\ \hline
                                        %   p                      gridsize             50        101      vone
\textbf{Baseline}                                   &                      &                    & 80   & 60    & 60    & 100  \\ \hline

\multirow{2}{6em}{\textbf{Randomized Image Shuffle}}             & 0.5                  &\multirow{2}{*}{}   & 50   & 25    & 75    & 100  \\ \cline{2-2} \cline{4-7} 
                                                    & 1                    &                    & 25   & 75 & 20 & 37.5    \\ \hline
\multirow{3}{6em}{\textbf{Grid Shuffle}}            &\multirow{3}{*}{}     & 4x4                & 20   & 40 & 0  & 80  \\ \cline{3-7} 
                                                    &                      & 8x8                & 40   & 40 & 40 & 90  \\ \cline{3-7} 
                                                    &                      & 16x16              & 20   & 0  & 40 & 100   \\ \hline
\multirow{6}{6em}{\textbf{Within Grid Shuffle}}     & \multirow{3}{*}{0.5} & 4x4                & 40   & 0  & 60 & 90  \\ \cline{3-7} 
                                                    &                      & 8x8                & 40   & 20 & 20 & 70  \\ \cline{3-7} 
                                                    &                      & 16x16              & 30   & 40 & 20 & 50  \\ \cline{2-7} 
                                                    & \multirow{3}{*}{1}   & 4x4                & 40   & 40 & 40 & 80   \\ \cline{3-7} 
                                                    &                      & 8x8                & 20   & 0  & 40 & 60  \\ \cline{3-7} 
                                                    &                      & 16x16              & 0    & 0  & 0  & 60   \\ \hline
\multirow{6}{6em}{\textbf{Local Grid Shuffle}} & \multirow{3}{*}{0.5} & 4x4                & 60   & 40 & 40 & 60   \\ \cline{3-7} 
                                                    &                      & 8x8                & 20   & 60 & 40 & 50  \\ \cline{3-7} 
                                                    &                      & 16x16              & 20   & 60 & 40 & 70  \\ \cline{2-7} 
                                                    & \multirow{3}{*}{1}   & 4x4                & 40   & 40 & 60 & 50  \\ \cline{3-7} 
                                                    &                      & 8x8                & 40   & 40 & 40 & 30  \\ \cline{3-7} 
                                                    &                      & 16x16              & 40   & 40 & 20 & 30   \\ \hline
\textbf{Color Flatten}                                 &                      &                    & 80   & 60 & 60 & 0  \\ \hline
\end{tabular}
\end{table*}
\end{center}

\begin{center}
\begin{table}
\caption{OLS for all data (baseline and transforms)}
\label{tab:ols_alldata}
\begin{tabular}{lclc}
\toprule
\textbf{Dep. Variable:}    &        y         & \textbf{  R-squared:         } &     0.139   \\
\textbf{Model:}            &       OLS        & \textbf{  Adj. R-squared:    } &     0.132   \\
\textbf{Method:}           &  Least Squares   & \textbf{  F-statistic:       } &     19.87   \\
\textbf{Date:}             & Sat, 07 May 2022 & \textbf{  Prob (F-statistic):} &  5.89e-12   \\
\textbf{Time:}             &     20:36:26     & \textbf{  Log-Likelihood:    } &   -241.54   \\
\textbf{No. Observations:} &         372      & \textbf{  AIC:               } &     491.1   \\
\textbf{Df Residuals:}     &         368      & \textbf{  BIC:               } &     506.8   \\
\textbf{Df Model:}         &           3      & \textbf{                     } &             \\
\bottomrule
\end{tabular}
\begin{tabular}{lcccccc}
               & \textbf{coef} & \textbf{std err} & \textbf{t} & \textbf{P$> |$t$|$} & \textbf{[0.025} & \textbf{0.975]}  \\
\midrule
\textbf{Humans} &       0.7957  &        0.048     &    16.478  &         0.000        &        0.701    &        0.891     \\
\textbf{VOneResNet50}    &      -0.4194  &        0.068     &    -6.141  &         0.000        &       -0.554    &       -0.285     \\
\textbf{ResNet101}    &      -0.4409  &        0.068     &    -6.456  &         0.000        &       -0.575    &       -0.307     \\
\textbf{ResNet50}    &      -0.4301  &        0.068     &    -6.298  &         0.000        &       -0.564    &       -0.296     \\
\bottomrule
\end{tabular}
\begin{tabular}{lclc}
\textbf{Omnibus:}       & 2791.922 & \textbf{  Durbin-Watson:     } &    2.020  \\
\textbf{Prob(Omnibus):} &   0.000  & \textbf{  Jarque-Bera (JB):  } &   33.509  \\
\textbf{Skew:}          &   0.228  & \textbf{  Prob(JB):          } & 5.29e-08  \\
\textbf{Kurtosis:}      &   1.602  & \textbf{  Cond. No.          } &     4.79  \\
\bottomrule
\end{tabular}

%\caption{OLS Regression Results}
\end{table}
\end{center}

% Warnings: \newline
%  [1] Standard Errors assume that the covariance matrix of the errors is correctly specified.

\begin{center}
\begin{table}
\caption{OLS for Baseline}
\label{tab:ols_baseline}
\begin{tabular}{lclc}
\toprule
\textbf{Dep. Variable:}    &        y         & \textbf{  R-squared:         } &     0.147   \\
\textbf{Model:}            &       OLS        & \textbf{  Adj. R-squared:    } &    -0.013   \\
\textbf{Method:}           &  Least Squares   & \textbf{  F-statistic:       } &    0.9167   \\
\textbf{Date:}             & Sat, 07 May 2022 & \textbf{  Prob (F-statistic):} &    0.455    \\
\textbf{Time:}             &     13:36:24     & \textbf{  Log-Likelihood:    } &   -10.053   \\
\textbf{No. Observations:} &          20      & \textbf{  AIC:               } &     28.11   \\
\textbf{Df Residuals:}     &          16      & \textbf{  BIC:               } &     32.09   \\
\textbf{Df Model:}         &           3      & \textbf{                     } &             \\
\bottomrule
\end{tabular}
\begin{tabular}{lcccccc}
               & \textbf{coef} & \textbf{std err} & \textbf{t} & \textbf{P$> |$t$|$} & \textbf{[0.025} & \textbf{0.975]}  \\
\midrule
\textbf{Humans} &       1.0000  &        0.200     &     5.000  &         0.000        &        0.576    &        1.424     \\
\textbf{VOneResNet50}    &      -0.4000  &        0.283     &    -1.414  &         0.176        &       -1.000    &        0.200     \\
\textbf{ResNet101}    &      -0.4000  &        0.283     &    -1.414  &         0.176        &       -1.000    &        0.200     \\
\textbf{ResNet50}    &      -0.2000  &        0.283     &    -0.707  &         0.490        &       -0.800    &        0.400     \\
\bottomrule

\end{tabular}
\begin{tabular}{lclc}
\textbf{Omnibus:}       &  3.241 & \textbf{  Durbin-Watson:     } &    2.525  \\
\textbf{Prob(Omnibus):} &  0.198 & \textbf{  Jarque-Bera (JB):  } &    2.513  \\
\textbf{Skew:}          & -0.750 & \textbf{  Prob(JB):          } &    0.285  \\
\textbf{Kurtosis:}      &  2.125 & \textbf{  Cond. No.          } &     4.79  \\
\bottomrule
\end{tabular}
\end{table}
%\caption{OLS Regression Results}
\end{center}
\clearpage
% Warnings: \newline
%  [1] Standard Errors assume that the covariance matrix of the errors is correctly specified.
\begin{center}
\begin{table}
\caption{OLS for Randomized Image Shuffle}
\label{tab:ols_randomshuffle}
\begin{tabular}{lclc}
\toprule
\textbf{Dep. Variable:}    &        y         & \textbf{  R-squared:         } &     0.048   \\
\textbf{Model:}            &       OLS        & \textbf{  Adj. R-squared:    } &    -0.054   \\
\textbf{Method:}           &  Least Squares   & \textbf{  F-statistic:       } &    0.4667   \\
\textbf{Date:}             & Sat, 07 May 2022 & \textbf{  Prob (F-statistic):} &    0.708    \\
\textbf{Time:}             &     13:45:31     & \textbf{  Log-Likelihood:    } &   -22.193   \\
\textbf{No. Observations:} &          32      & \textbf{  AIC:               } &     52.39   \\
\textbf{Df Residuals:}     &          28      & \textbf{  BIC:               } &     58.25   \\
\textbf{Df Model:}         &           3      & \textbf{                     } &             \\
\bottomrule
\end{tabular}
\begin{tabular}{lcccccc}
               & \textbf{coef} & \textbf{std err} & \textbf{t} & \textbf{P$> |$t$|$} & \textbf{[0.025} & \textbf{0.975]}  \\
\midrule
\textbf{Humans} &       0.7500  &        0.183     &     4.099  &         0.000        &        0.375    &        1.125     \\
\textbf{VOneResNet50}    &      -0.2500  &        0.259     &    -0.966  &         0.342        &       -0.780    &        0.280     \\
\textbf{ResNet101}    &      -0.2500  &        0.259     &    -0.966  &         0.342        &       -0.780    &        0.280     \\
\textbf{ResNet50}    &      -0.2500  &        0.259     &    -0.966  &         0.342        &       -0.780    &        0.280     \\
\bottomrule
\end{tabular}
\begin{tabular}{lclc}
\textbf{Omnibus:}       & 60.696 & \textbf{  Durbin-Watson:     } &    2.275  \\
\textbf{Prob(Omnibus):} &  0.000 & \textbf{  Jarque-Bera (JB):  } &    4.421  \\
\textbf{Skew:}          & -0.207 & \textbf{  Prob(JB):          } &    0.110  \\
\textbf{Kurtosis:}      &  1.227 & \textbf{  Cond. No.          } &     4.79  \\
\bottomrule
\end{tabular}
%\caption{OLS Regression Results}
\end{table}
\end{center}

\begin{center}
\begin{table}
\caption{OLS for Grid Shuffle (all parameters)}
\label{tab:ols_gridshuffle}
\begin{tabular}{lclc}
\toprule
\textbf{Dep. Variable:}    &        y         & \textbf{  R-squared:         } &     0.407   \\
\textbf{Model:}            &       OLS        & \textbf{  Adj. R-squared:    } &     0.376   \\
\textbf{Method:}           &  Least Squares   & \textbf{  F-statistic:       } &     12.83   \\
\textbf{Date:}             & Sat, 07 May 2022 & \textbf{  Prob (F-statistic):} &  1.72e-06   \\
\textbf{Time:}             &     13:40:05     & \textbf{  Log-Likelihood:    } &   -27.549   \\
\textbf{No. Observations:} &          60      & \textbf{  AIC:               } &     63.10   \\
\textbf{Df Residuals:}     &          56      & \textbf{  BIC:               } &     71.47   \\
\textbf{Df Model:}         &           3      & \textbf{                     } &             \\
\bottomrule
\end{tabular}
\begin{tabular}{lcccccc}
               & \textbf{coef} & \textbf{std err} & \textbf{t} & \textbf{P$> |$t$|$} & \textbf{[0.025} & \textbf{0.975]}  \\
\midrule
\textbf{Humans} &       1.0000  &        0.102     &     9.770  &         0.000        &        0.795    &        1.205     \\
\textbf{VOneResNet50}    &      -0.7333  &        0.145     &    -5.066  &         0.000        &       -1.023    &       -0.443     \\
\textbf{ResNet101}    &      -0.7333  &        0.145     &    -5.066  &         0.000        &       -1.023    &       -0.443     \\
\textbf{ResNet50}    &      -0.7333  &        0.145     &    -5.066  &         0.000        &       -1.023    &       -0.443     \\
\bottomrule
\end{tabular}
\begin{tabular}{lclc}
\textbf{Omnibus:}       & 12.547 & \textbf{  Durbin-Watson:     } &    2.281  \\
\textbf{Prob(Omnibus):} &  0.002 & \textbf{  Jarque-Bera (JB):  } &   14.931  \\
\textbf{Skew:}          &  1.219 & \textbf{  Prob(JB):          } & 0.000572  \\
\textbf{Kurtosis:}      &  2.818 & \textbf{  Cond. No.          } &     4.79  \\
\bottomrule
\end{tabular}
\end{table}
%\caption{OLS Regression Results}
\end{center}

% Warnings: \newline
%  [1] Standard Errors assume that the covariance matrix of the errors is correctly specified.
\clearpage
\begin{center}
\begin{table}
\caption{OLS for Within Grid Shuffle}
\label{tab:ols_withingridshuffle}
\begin{tabular}{lclc}
\toprule
\textbf{Dep. Variable:}    &        y         & \textbf{  R-squared:         } &     0.325   \\
\textbf{Model:}            &       OLS        & \textbf{  Adj. R-squared:    } &     0.308   \\
\textbf{Method:}           &  Least Squares   & \textbf{  F-statistic:       } &     18.62   \\
\textbf{Date:}             & Sat, 07 May 2022 & \textbf{  Prob (F-statistic):} &  6.32e-10   \\
\textbf{Time:}             &     13:46:55     & \textbf{  Log-Likelihood:    } &   -60.629   \\
\textbf{No. Observations:} &         120      & \textbf{  AIC:               } &     129.3   \\
\textbf{Df Residuals:}     &         116      & \textbf{  BIC:               } &     140.4   \\
\textbf{Df Model:}         &           3      & \textbf{                     } &             \\
\bottomrule
\end{tabular}
\begin{tabular}{lcccccc}
               & \textbf{coef} & \textbf{std err} & \textbf{t} & \textbf{P$> |$t$|$} & \textbf{[0.025} & \textbf{0.975]}  \\
\midrule
\textbf{Humans} &       0.8667  &        0.074     &    11.638  &         0.000        &        0.719    &        1.014     \\
\textbf{VOneResNet50}    &      -0.5667  &        0.105     &    -5.381  &         0.000        &       -0.775    &       -0.358     \\
\textbf{ResNet101}    &      -0.7000  &        0.105     &    -6.647  &         0.000        &       -0.909    &       -0.491     \\
\textbf{ResNet50}    &      -0.6333  &        0.105     &    -6.014  &         0.000        &       -0.842    &       -0.425     \\
\bottomrule
\end{tabular}
\begin{tabular}{lclc}
\textbf{Omnibus:}       &  9.985 & \textbf{  Durbin-Watson:     } &    2.029  \\
\textbf{Prob(Omnibus):} &  0.007 & \textbf{  Jarque-Bera (JB):  } &   10.557  \\
\textbf{Skew:}          &  0.726 & \textbf{  Prob(JB):          } &  0.00510  \\
\textbf{Kurtosis:}      &  3.065 & \textbf{  Cond. No.          } &     4.79  \\
\bottomrule
\end{tabular}
%\caption{OLS Regression Results}
\end{table}
\end{center}

  \begin{center}
 \begin{table}
 \caption{OLS for Local Grid Shuffle}
 \label{tab:ols_localshuffle}
\begin{tabular}{lclc}
\toprule
\textbf{Dep. Variable:}    &        y         & \textbf{  R-squared:         } &     0.083   \\
\textbf{Model:}            &       OLS        & \textbf{  Adj. R-squared:    } &     0.059   \\
\textbf{Method:}           &  Least Squares   & \textbf{  F-statistic:       } &     3.503   \\
\textbf{Date:}             & Sat, 07 May 2022 & \textbf{  Prob (F-statistic):} &   0.0177    \\
\textbf{Time:}             &     13:44:13     & \textbf{  Log-Likelihood:    } &   -81.874   \\
\textbf{No. Observations:} &         120      & \textbf{  AIC:               } &     171.7   \\
\textbf{Df Residuals:}     &         116      & \textbf{  BIC:               } &     182.9   \\
\textbf{Df Model:}         &           3      & \textbf{                     } &             \\
\bottomrule
\end{tabular}
\begin{tabular}{lcccccc}
               & \textbf{coef} & \textbf{std err} & \textbf{t} & \textbf{P$> |$t$|$} & \textbf{[0.025} & \textbf{0.975]}  \\
\midrule
\textbf{Humans} &       0.7333  &        0.089     &     8.249  &         0.000        &        0.557    &        0.909     \\
\textbf{VOneResNet50}    &      -0.3333  &        0.126     &    -2.651  &         0.009        &       -0.582    &       -0.084     \\
\textbf{ResNet101}    &      -0.2667  &        0.126     &    -2.121  &         0.036        &       -0.516    &       -0.018     \\
\textbf{ResNet50}    &      -0.3667  &        0.126     &    -2.917  &         0.004        &       -0.616    &       -0.118     \\
\bottomrule
\end{tabular}
\begin{tabular}{lclc}
\textbf{Omnibus:}       & 1722.081 & \textbf{  Durbin-Watson:     } &    2.362  \\
\textbf{Prob(Omnibus):} &   0.000  & \textbf{  Jarque-Bera (JB):  } &   13.905  \\
\textbf{Skew:}          &   0.080  & \textbf{  Prob(JB):          } & 0.000956  \\
\textbf{Kurtosis:}      &   1.340  & \textbf{  Cond. No.          } &     4.79  \\
\bottomrule
\end{tabular}
%\caption{OLS Regression Results}
\end{table}
\end{center}

\clearpage
 \begin{center}
 \begin{table}
 \caption{OLS for Color Flatten}
 \label{tab:ols_colordist}
\begin{tabular}{lclc}
\toprule
\textbf{Dep. Variable:}    &        y         & \textbf{  R-squared:         } &     0.360   \\
\textbf{Model:}            &       OLS        & \textbf{  Adj. R-squared:    } &     0.240   \\
\textbf{Method:}           &  Least Squares   & \textbf{  F-statistic:       } &     3.000   \\
\textbf{Date:}             & Sat, 07 May 2022 & \textbf{  Prob (F-statistic):} &   0.0615    \\
\textbf{Time:}             &     13:41:55     & \textbf{  Log-Likelihood:    } &   -10.053   \\
\textbf{No. Observations:} &          20      & \textbf{  AIC:               } &     28.11   \\
\textbf{Df Residuals:}     &          16      & \textbf{  BIC:               } &     32.09   \\
\textbf{Df Model:}         &           3      & \textbf{                     } &             \\
\bottomrule
\end{tabular}
\begin{tabular}{lcccccc}
               & \textbf{coef} & \textbf{std err} & \textbf{t} & \textbf{P$> |$t$|$} & \textbf{[0.025} & \textbf{0.975]}  \\
\midrule
\textbf{Humans} &    5.392e-16  &        0.200     &   2.7e-15  &         1.000        &       -0.424    &        0.424     \\
\textbf{VOneResNet50}    &       0.6000  &        0.283     &     2.121  &         0.050        &        0.000    &        1.200     \\
\textbf{ResNet101}    &       0.6000  &        0.283     &     2.121  &         0.050        &        0.000    &        1.200     \\
\textbf{ResNet50}    &       0.8000  &        0.283     &     2.828  &         0.012        &        0.200    &        1.400     \\
\bottomrule
\end{tabular}
\begin{tabular}{lclc}
\textbf{Omnibus:}       &  3.241 & \textbf{  Durbin-Watson:     } &    1.900  \\
\textbf{Prob(Omnibus):} &  0.198 & \textbf{  Jarque-Bera (JB):  } &    2.513  \\
\textbf{Skew:}          & -0.750 & \textbf{  Prob(JB):          } &    0.285  \\
\textbf{Kurtosis:}      &  2.125 & \textbf{  Cond. No.          } &     4.79  \\
\bottomrule
\end{tabular}
%\caption{OLS Regression Results}
\end{table}
\end{center}

% Warnings: \newline
%  [1] Standard Errors assume that the covariance matrix of the errors is correctly specified.

\begin{center}
\begin{table}
\caption{OLS for all transforms with $16 \times 16$ grid}
\label{tab:ols_16x16}
\begin{tabular}{lclc}
\toprule
\textbf{Dep. Variable:}    &        y         & \textbf{  R-squared:         } &     0.253   \\
\textbf{Model:}            &       OLS        & \textbf{  Adj. R-squared:    } &     0.230   \\
\textbf{Method:}           &  Least Squares   & \textbf{  F-statistic:       } &     10.84   \\
\textbf{Date:}             & Sat, 07 May 2022 & \textbf{  Prob (F-statistic):} &  3.39e-06   \\
\textbf{Time:}             &     20:32:05     & \textbf{  Log-Likelihood:    } &   -55.030   \\
\textbf{No. Observations:} &         100      & \textbf{  AIC:               } &     118.1   \\
\textbf{Df Residuals:}     &          96      & \textbf{  BIC:               } &     128.5   \\
\textbf{Df Model:}         &           3      & \textbf{                     } &             \\
\bottomrule
\end{tabular}
\begin{tabular}{lcccccc}
               & \textbf{coef} & \textbf{std err} & \textbf{t} & \textbf{P$> |$t$|$} & \textbf{[0.025} & \textbf{0.975]}  \\
\midrule
\textbf{Humans} &       0.8000  &        0.086     &     9.342  &         0.000        &        0.630    &        0.970     \\
\textbf{VOneResNet50}    &      -0.5600  &        0.121     &    -4.624  &         0.000        &       -0.800    &       -0.320     \\
\textbf{ResNet101}    &      -0.5200  &        0.121     &    -4.294  &         0.000        &       -0.760    &       -0.280     \\
\textbf{ResNet50}    &      -0.6000  &        0.121     &    -4.954  &         0.000        &       -0.840    &       -0.360     \\
\bottomrule
\end{tabular}
\begin{tabular}{lclc}
\textbf{Omnibus:}       &  6.706 & \textbf{  Durbin-Watson:     } &    2.167  \\
\textbf{Prob(Omnibus):} &  0.035 & \textbf{  Jarque-Bera (JB):  } &    6.940  \\
\textbf{Skew:}          &  0.621 & \textbf{  Prob(JB):          } &   0.0311  \\
\textbf{Kurtosis:}      &  2.653 & \textbf{  Cond. No.          } &     4.79  \\
\bottomrule
\end{tabular}
%\caption{OLS Regression Results}
\end{table}
\end{center}

% Warnings: \newline
%  [1] Standard Errors assume that the covariance matrix of the errors is correctly specified.

\end{document}